\documentclass[10pt,conference]{IEEEtran}
\IEEEoverridecommandlockouts
\usepackage{epsfig}
\usepackage{verbatim}
\usepackage{balance}
\usepackage{amsmath}
\usepackage{cite}
\usepackage{float}
\usepackage{algorithm}
\usepackage{amssymb}
\usepackage[english]{babel}
\usepackage{array}
\usepackage{comment}
\usepackage{multirow}
\usepackage{enumerate}
\usepackage{color}
\usepackage{xcolor}
\usepackage{listings}
\usepackage{graphicx}
\usepackage{epstopdf}
\usepackage{graphics}
\usepackage{subfigure}
\usepackage{hyperref}
\usepackage{enumitem}
\usepackage{textcomp}
\usepackage{caption}
\usepackage{tikz}
\usepackage{textcomp}
\usepackage[doipre={DOI:~}]{uri}
\usepackage{lipsum}

\usepackage{tikz}
\newcommand*\circled[1]{\tikz[baseline=(char.base)]{
            \node[shape=circle,draw,inner sep=1pt] (char) {#1};}}
            
\usepackage{tabularx}
\newenvironment{conditions*}
{\par\vspace{\abovedisplayskip}\noindent
	\tabularx{\columnwidth}{>{$}l<{$} @{${}={}$} >{\raggedright\arraybackslash}X}}
{\endtabularx\par\vspace{\belowdisplayskip}}

\newcommand\copyrighttext{%
  \footnotesize \textcopyright 2022 IEEE. Personal use of this material is permitted.  Permission from IEEE must be obtained for all other uses, in any current or future media, including reprinting/republishing this material for advertising or promotional purposes, creating new collective works, for resale or redistribution to servers or lists, or reuse of any copyrighted component of this work in other works. 
  
  Accepted as a conference paper at the 2022 IEEE International Conference on Omni-Layer Intelligent Systems (COINS).}
\newcommand{\copyrightnotice}{%
\begin{tikzpicture}[remember picture,overlay]
\node[anchor=south,yshift=10pt] at (current page.south) {\fbox{\parbox{\dimexpr\textwidth-\fboxsep-\fboxrule\relax}{\copyrighttext}}};
\end{tikzpicture}%
}

\begin{document}

\title{A Machine Learning-based Digital Twin for Electric Vehicle Battery Modeling}

\author{
\IEEEauthorblockN{Khaled Sidahmed Sidahmed Alamin$^1$, Yukai Chen$^1$, Enrico Macii$^2$, Massimo Poncino$^1$, Sara Vinco$^1$}
\IEEEauthorblockA{
\IEEEauthorrefmark{1}Department of Control and Computer Engineering, Politecnico di Torino, Turin, Italy \\
\IEEEauthorrefmark{2}Interuniversity Department of Regional and Urban Studies and Planning, Politecnico di Torino, Turin, Italy}%
\IEEEauthorblockA{Email: name.surname@polito.it}%
}


\maketitle

\copyrightnotice

\begin{abstract}
The widespread adoption of EVs is limited by their reliance on batteries with presently low energy and power densities compared to liquid fuels and are subject to aging and performance deterioration over time. For this reason, monitoring the battery state of charge and state of health during the EV lifetime is a very relevant problem. This work proposes the structure of a battery digital twin designed to reflect battery dynamics at the run time accurately. To ensure a high degree of correctness concerning non-linear phenomena, the digital twin relies on data-driven models trained on traces of battery evolution over time: a state of health model, repeatedly executed to estimate the degradation of maximum battery capacity, and a state of charge model, retrained periodically to reflect the impact of aging. The proposed digital twin structure will be exemplified on a public dataset to motivate its adoption and prove its effectiveness, with a high degree of accuracy and inference and retraining times compatible with onboard execution.         
\end{abstract}

\section{Introduction}
Since General Motors sold the first Electric Vehicle (EV) in 1996, the market of EVs has experienced explosive growth, and in the past five years, global electric car sales have more than quadrupled \cite{EVmarket}. This dramatic growth is an effect of both environmental and economic considerations, that lead governments to lower CO$_2$ emissions standards and to implement increased subsidy schemes for EVs as part of stimulus packages to counter the effects of the pandemic \cite{EVmarket2}. However, the effectiveness of EVs is limited by their reliance on batteries that have presently low energy and power densities compared to liquid fuels and are subject to aging and performance deterioration over time \cite{techroadmap}. 

\begin{figure}[!htbp]
\centering
\includegraphics[width=0.9\linewidth]{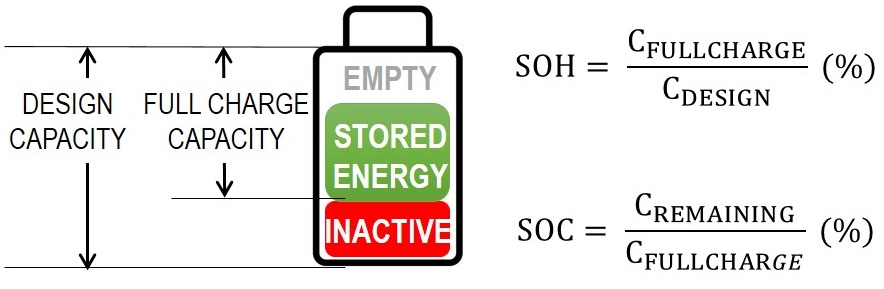}
\caption{
  \label{fig:intro}
  Representation of battery SOH ans SOC metrics: SoH quantifies battery capacity degradation and impacts on the estimation of SOC.
}
	\vspace{-0.6\baselineskip}
\end{figure}

To partially overcome these limitations, EVs are usually powered by Lithium-ion batteries that have high power and energy density and long cycle life compared to other chemistry compositions \cite{techroadmap}. Nonetheless, battery management is critical to enhancing safety, reliability, and performance. In particular, accurate estimation of battery \emph{State of Charge (SOC)} and \emph{State of Health (SOH)} is crucial to deliver essential information about battery charge and aging level, which can be used to perform maintenance or adapt management strategies to improve performance and reliability (Figure \ref{fig:intro}) \cite{LI2020101557}. 

Unfortunately, both metrics must reflect complex non-linear battery dynamics, and modern Battery Management Systems (BMS) require knowledge of historical data of battery usage that can not be quickly processed onboard in real-time. For this reason, \emph{battery digital twins} have been introduced, aiming at building an accurate run time model of the SOC and SOH metrics to provide accurate battery capacity degradation assessment and relevant information to the BMS \cite{LI2020101557}. This is achieved through a combination of Internet of Things communication protocols, cloud/edge computation, and machine learning applications that jointly achieve an exponential increase in data storage and computation capability \cite{LI2020101557,CHEN2013184}. 

The main ingredient of a digital twin is a model of the process/system under monitoring, in this case, the relevant battery SOH and SOC dynamics: the model is used as a reference for the process under analysis to detect any misalignment or unexpected behavior, and to foresee the future evolution. In the literature, many battery models have been proposed, ranging from electrochemical models to circuit-equivalent models~\cite{chen2017circuit}. Machine learning techniques are becoming increasingly popular for battery SOC and SOH estimation~\cite{vidal2020machine}. Finding a balance between modeling accuracy, model responsiveness, sensitivity to battery evolution, and the architectural constraint imposed by the EV scenario is far from trivial. Additionally, as clarified by Figure \ref{fig:intro}, SOH and SOC are tightly interleaved: charge and discharge cycles affect battery SOH, which reduces maximum total charge capacity and the corresponding estimation of battery SOC. This implies that the SOH and SOC models can not be built independently and that the SOC dynamics must be aware of the status of the SOH metrics. 

This paper discusses the challenges posed by the construction of a battery digital twin, and \emph{proposes a custom architecture of models for SOC and SOH to propose a viable solution for an accurate estimation of battery dynamics and aging effects}.  
In detail: 
\begin{itemize}[leftmargin=*]
    \item the digital twin provides a model of SOH built on historical data, that is trained once on data of a complete charge cycle; 
    \item the model of the SOC is heavily affected by aging: for this reason, the digital twin features a model of SOC that is retrained periodically on the cloud on data uploaded from the EV. The model is then run either on the cloud (to have higher performance) or on the EV (to provide the accurate estimation to the onboard BMS); 
    \item the paper discusses the edge/cloud architecture, even if discussing the details of the digital twin infrastructure (edge/cloud, protocols, etc.) is out of the scope of this paper and will be part of future work; 
    \item we apply the proposed solution to model to a public dataset provided by NASA Ames Research Centre \cite{dataset}, to motivate its adoption, prove the effectiveness of the proposed solution in modeling Lithium-ion batteries, and provide implementations of the SOC and SOH models.  
\end{itemize}

The paper is organized as follows. Section \ref{sec:background} provides the necessary background on battery dynamics and digital twins. Section \ref{sec:dtmodels} presents the proposed solution, with a strong focus on the digital twin architecture and on how to reproduce effectively the interconnection between SOH and SOC. Section \ref{sec:nasaexperiments} applies the digital twin concept to the NASA Ames Research Centre dataset, and finally Section \ref{sec:conclusions} draws our concluding remarks.

\section{Background}\label{sec:background}
\subsection{Battery Dynamics}
Lithium-ion batteries store and supply electric power based on the movement of Li-ions between anode and cathode. Some irreversible reactions happening during the repeated charges and discharges cycles (such as Li planting and growth of solid electrolyte interface) cause performance degradation over time, generating issues in terms of reduced available capacity and/or increased internal resistance \cite{ZHANG2018288}. 

\emph{State of Charge (SOC)} estimation is a relevant metric that measures the percentage of remaining capacity relative to the maximum capacity of the battery. To provide robust estimates, equivalent circuit models are used that achieve good approximation but require detailed physical knowledge of Li-ion batteries. The complementary approach uses data-based models that reconstruct the evolution of battery SOC given curves of, e.g., current, voltage and temperature. Many solutions have been proposed in the literature, using e.g., support vector machines \cite{6423937,YAO2021118866} and neural networks \cite{CHANG2013603,HE2014783,XU201233,9028181}. The main issue of these approaches is that \emph{they build a once-for-all model of SOC that is applied throughout the battery lifetime, thus missing the impact of aging on the internal evolution of battery.} This is especially critical as one of the major impacts of aging is a reduction in battery available maximum capacity, which impacts battery SOC calculation. 

Tracking age-dependent changes in battery dynamics is thus necessary to maintain accurate estimates of the remaining lifetime of the battery. In this perspective, \emph{State of Health (SOH)} is a metric that measures the condition of a battery compared to its ideal conditions and is typically estimated as the ratio of the aged and rated capacity of the battery. SOH is not an observable quantity and is non-linearly dependent on several factors, including charge and discharge profiles, average SOC, and ambient temperature \cite{en15031185}. Given the difficulty in modeling the underlying electrochemical processes, most approaches estimate SOH models with data-driven methods, ranging from genetic algorithms to learning- and neural network-based algorithms~\cite{CHEN2013184,SUI2021117346}. 

\subsection{Digital Twins for Battery Management}
A digital twin is a virtual representation that serves as the real-time digital counterpart of a physical object or process. In the case of batteries, digital twins are now being used to complement BMS with high computation power and data storage capability, reliability, and accuracy \cite{WU2020100016}. Works at state of the art differ in terms of adopted battery model. Circuit-equivalent models (e.g., the Thevenin model) are complicated to characterize and may even require experimental measurements to ensure a reasonable degree of correctness \cite{LI2020101557}. Algorithmic approaches such as genetic algorithms and particle swarm optimization are characterized by low convergence and are computationally expensive, thus not allowing to provide estimations at run time during EV lifetime \cite{CHEN2013184}. Data analytic strategies (like machine learning, support vector machines, and neural networks) overcome these issues by self-learning parameters from historical data. However, they are trained at the beginning of battery lifetime and thus tend to diverge from the actual aged battery along its lifetime \cite{8240689,6423937,YAO2021118866,CHANG2013603,HE2014783,XU201233,9028181, QU2020113857}.  
In this direction, an interesting strategy has been proposed by \cite{LI2020101557}, which proposes to periodically influence the SOC model with updated SOH information, to improve prediction accuracy. This solution improves prediction accuracy by better reflecting the impact of aging on internal battery dynamics. However, this work adopts the classic H-infinity filter and particle swarm optimization techniques for SOC and SOH estimation, and they require a cumbersome parameter identification process every time SOC and SOH estimation is performed. Moreover, the H-infinity filter for SOC estimation is fixed all the time, only leveraging the SOH value to affect the adaptive process in the filter.

\section{A Digital Twin for Battery Monitoring}\label{sec:dtmodels}
The proposed digital twin of the EV battery storage system focuses on SOC and SOH estimation as the most significant parameters that affect EV's driving experience and battery life. 

\subsection{Models for Battery SOC and SOH}

\begin{figure}[!htbp]
	\vspace{-1\baselineskip}
	\begin{center}
		\includegraphics[width=0.9\linewidth]{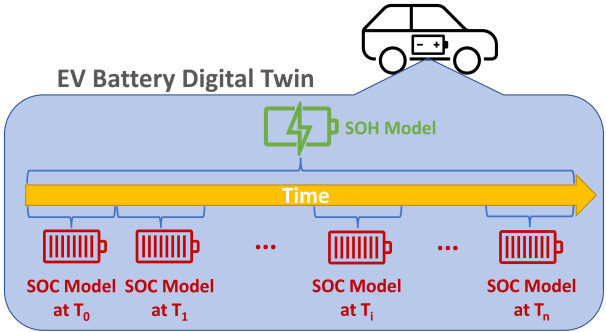}
		\caption{Conceptual diagram of the EV battery digital twin: the arrow is battery lifetime, the icon on top represents the SOH data-driven model, while the icons below the arrow represent the SOC data-driven models that are retrained periodically during battery lifetime.}
		\vspace{-1\baselineskip}
		\label{fig:concept} 
	\end{center}
\end{figure}

Figure~\ref{fig:concept} shows the conceptual diagram of our proposed model. The horizontal arrow represents battery lifetime, from nominal conditions (SOH = 100\%) to the point when the battery is considered no longer usable (SOH = 80\%). The battery icons represent two different classes of models: 

\begin{itemize}[leftmargin=*]
\item A model of SOH (on top of the arrow), built once for all on historical data and run throughout the whole lifetime of the battery; 
\item A model of SOC (below the arrow) that is periodically updated to reflect the impact of SOH on the estimation of full battery charge and thus of SOC. The multiple icons reflect the fact that different SOC models are used by the digital twin to estimate the battery SOC in different battery operation time intervals: the model at $T_i$ shown in Figure~\ref{fig:concept} represents the latest retrained SOC model adapted to the battery conditions at the $i_{th}$ time interval. 
\end{itemize}

Both models are data-driven in the sense that they are built based on data traces. It is important to note that any machine learning-based solution can be used to build the battery SOC and SOH models~\cite{vidal2020machine}. The battery SOH model is built once for all from historical data of an aging battery of the same type as the one installed on the vehicle. The battery SOC model is initially built on historical data of nominal battery conditions and is then retrained on data sampled from the vehicle during its operation. Two mechanisms can trigger the update of the SOC model: a fixed period or a decrease of SOH by a specific value. 

The reason for such a static SOH model combined with a dynamic SOC model design is that the SOH of the battery from 100\% to 80\% represents the aging of the total available capacitance over the entire operating life cycle of the battery, where the only relationship that changes is that the total available capacitance decreases with the time of use. On the other hand, the SOC model should be constantly updated because the discharge characteristics of the battery are different under different SOH conditions. More specifically, the battery impedance and battery maximum capacity change as SOH decreases; furthermore, the relationship between SOC and battery open-circuit voltage during discharge differs under different SOH conditions~\cite{en14154506}. 

\subsection{Digital Twin Architecture and Operation}

\begin{figure}[!htbp]
		\vspace{-1\baselineskip}
	\begin{center}
		\includegraphics[width=0.96\linewidth]{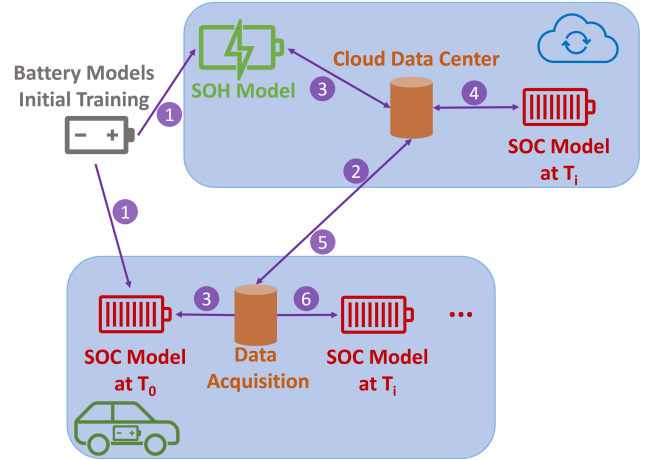}
		\vspace{-1\baselineskip}
		\caption{Methodology to implement EV battery digital twin.}
		\vspace{-1\baselineskip}
		\label{fig:method} 
	\end{center}
\end{figure}

Figure~\ref{fig:method} illustrates more details about how to construct our proposed digital twin and its working mechanism. The diagram shows that our proposed digital twin has two main components, the cloud side and the EV edge side: 

\begin{itemize}
    \item The cloud is used to deploy and run the SOH model, and to periodically retrain the SOC model; 
    \item The edge node on the vehicle is used to collect and periodically upload data to the cloud for SOC model retraining, and of running the updated SOC models; this choice allows the vehicle BMS to be constantly aware of batter SOC estimation. 
\end{itemize} 

The specific steps related to the number shown in the Figure~\ref{fig:method} for constructing our proposed digital twin and how it works during operation are described below.

\begin{itemize}[leftmargin=*]
    \item Step \circled{1}: offline training of the SOH model using machine learning techniques on historical measurements of an aged battery; training of the initial SOC model by using a discharge trace of a brand new battery (at nominal battery conditions);
    \item Step \circled{2}: the EV continuously collects online charge and discharge measurements data (i.e., samples of voltage, current, temperature, relative discharge and charge time of the battery) and uploads them to the cloud; 
    \item Step \circled{3}: the SOC model is executed on edge in the EV to compute SOC estimation, and 
    the SOH model deployed in the cloud leverages the latest uploaded battery measurements to compute the updated SOH;
    \item Step \circled{4}: when the pre-defined conditions for retraining the SOC model are met (e.g., SOH reduced by 1\%, or fixed time interval elapsed), the cloud uses the updated measurement data of the battery to retrain the SOC model; 
    \item Step \circled{5}: the retrained SOC model is transferred from the cloud to the EV edge node;
    \item Step \circled{6}: the EV edge node updates the deployed SOC model and goes back to step \circled{2} to execute the following steps repeatedly.
\end{itemize}

\section{Experiments}\label{sec:nasaexperiments} 
The goal of the proposed solution is not to propose a complete description of the digital twin, but to rather discuss the effectiveness of the proposed SOC-SOH models organization in the context of battery digital twin construction. For this reason, determining the optimal machine learning algorithms or the optimal cloud/edge architecture is beyond the scope of this work. This section applies the proposed SOC-SOH model structure to estimate its effectiveness in reflecting battery dynamics and the impact of aging. 

\subsection{Experiments Setup}

\subsubsection{Dataset Description}
The dataset used in this work is a NASA Ames Prognostics Center of Excellence (PCoE) dataset containing traces of 28 lithium-cobalt-oxide (LCO) 18650 batteries with 2.1Ah nominal capacity and 4.2V rated voltage that were continuously operated using a sequence of charging and discharging currents at different temperatures emulating random real-world battery usage~\cite{bole2014adaptation}. 
The dataset contains information about battery current, voltage, temperature, relative time in each reference discharge and charge cycle, and time during the whole measurements are tracked for both reference cycles and randomized cycles (also called Random Walk, RW). Reference charging and discharging cycles were performed periodically to provide reference points for computing the battery's total available capacity, which allows deriving the battery SOH in different moments of battery lifetime.

\subsubsection{Data Processing}
In this work, we selected 10 out of the 28 battery cells in the dataset, discarding the cells with malformed data (e.g., too low operating temperature or nominal capacity). 
We keep the reference discharge data in the measurements and remove the random walk portions. For example, after removing all the random walk cycles, RW9 data contains 79 reference discharge cycles, and RW13 has 22 reference discharge cycles, spanning the entire lifetime of battery measurements. We adopt all reference discharge data to construct our dataset in this work because we can use it to build both the SOH and SOC models and retrain the SOC model when SOH decreases.

Within each reference discharge cycle, we conduct the Coulomb Counting method to calculate the residual capacity corresponding to each time step in the discharge phase and compute the total available discharge capacity at the end of the discharge cycle. Then we convert the residual capacity during one reference discharge to the corresponding value of SOC and use each reference discharge's total available capacity to derive the SOH.
Notice that the SOH value in one discharge cycle keeps constant; it only updates from \textcolor{black}{discharge cycle to discharge cycle}.

\subsubsection{Motivation of The Adopted Digital Twin Strategy}

\begin{figure}[!htbp]
	\vspace{-0.6\baselineskip}
	\begin{center}
		\includegraphics[width=0.9\linewidth]{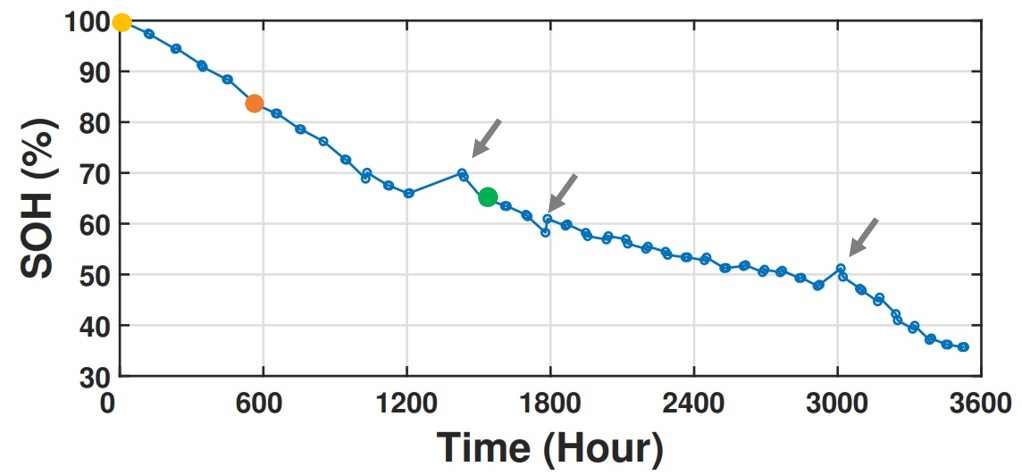}
		\caption{SOH degradation profile extracted from the RW9 dataset. Arrows points where the monotonically decreasing nature of SOH is not reflected due to measurement errors.}
		\vspace{-1\baselineskip}
		\label{fig:soh_dirty} 
	\end{center}
\end{figure}

Figure~\ref{fig:soh_dirty} indicates the extracted SOH decreasing profile during the whole lifetime of battery cell RW9, and each dot on the curve corresponds to a value of SOH computer from a reference discharge for RW9 provided in the dataset. 

\begin{figure}[!htbp]
	\vspace{-1\baselineskip}
	\begin{center}
		\includegraphics[width=0.9\linewidth]{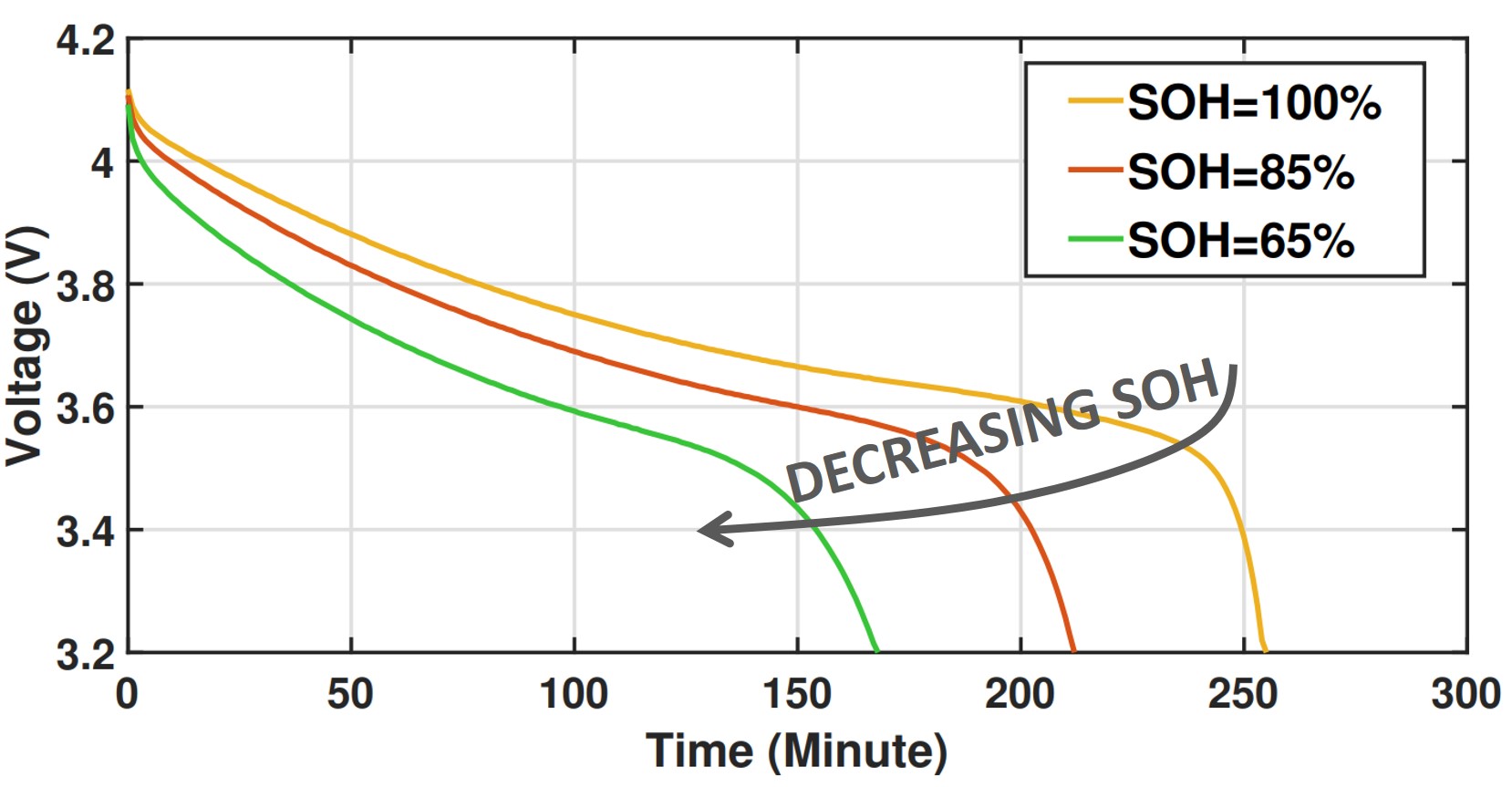}
        \caption{Voltage profile of reference discharge at different SOH values: a lower value of SOH leads to a reduction of overall discharge time (due to the reduction of the total available capacity) and to a change in the downward shape of the overall voltage discharge profile.}
		\label{fig:soh_voltage} 
	\end{center}
	\vspace{-1\baselineskip}
\end{figure}

Each dot in Figure~\ref{fig:soh_dirty} corresponds to a different voltage discharge curve, and different SOH values imply significant differences in the discharge curve. 
Figure~\ref{fig:soh_voltage} exemplifies the discharge voltage curve for the colored points in Figure \ref{fig:soh_dirty} (corresponding to SOH 100\%, 85\% and 65\%). The curves reflect that when SOH decreases, the overall discharge time is shortened due to the reduction of the total available capacity of the battery. Note that there is also a change in the downward shape of the overall voltage discharge profile. All these changes support the idea that it is necessary to retrain the SOC model, and that a SOC model can not be accurate throughout the whole battery lifetime.

\subsubsection{Data Cleaning}

After computing the SOH value for each reference discharge cycle, it was possible to identify some irregular cycles break in the monotonic decreasing nature of SOH due to the measurement errors (e.g., the points highlighted by the arrows in Figure~\ref{fig:soh_dirty}). We removed the erroneous samples from the data traces for all discharge cycles. For instance, over 79 reference discharge cycles, RW9 has nine samples with these erroneous behaviors, and they have been removed from the SOH profile to build the final dataset.

Ultimately, our dataset consists of the following columns, current, voltage, temperature, relative time, SOC, and SOH; the first four are used as inputs to train the models in our proposed digital twin, and the last two are used as outputs of the models.


\subsection{Machine Learning Techniques Used in This Work}\label{sec:models}

In this work, to perform the SOH and SOC estimation, we have experimented with different machine learning-based models. Choosing the optimal model configuration (models types and architectures) is beyond the scope of this work; the focus of this work is to clarify the architecture of our proposed digital twin and its operation mechanism, and in our future work, we will explore the optimal models' configuration based on different design requirements. Machine learning methods adopted in this work include:

\paragraph{Random Forest (RF)} A RF comprises many decision trees: when performing the classification task, new input samples enter, and each decision tree in the forest classifies them separately; the classification results chosen by most decision trees is taken as the final result~\cite{breiman2001random}.

\paragraph{Light Gradient Boosting (LGB)} LGB is a gradient learning system based on decision trees and the boosting concept~\cite{ke2017lightgbm}. It employs a leaf-wise growth approach with depth limitations and uses histogram-based techniques to speed up the training process and decrease memory usage.


\paragraph{Deep Neural Network (DNN)} The DNN adopted in our work is a feed-forward artificial neural network with more than one layer of hidden units between its inputs and outputs layers. Each hidden unit typically employs the logistic function to map its total input from the layer below to the scalar state that it sends to the layer above~\cite{6296526}. The neural network architecture used in this experiment is based on input data connected with seven hidden layers and the output layer with a linear activation (since it is a regression problem) function. 

During the training phase of all the methods mentioned above, the verification method uses k-fold cross-validation. The SOH and SOC estimation results are compared based on different models' accuracy, training time, and inference time. 

\subsection{SOH Estimation}

\subsubsection{Feature Selection}
There is a strong relationship between the relative time discharge time and the corresponding voltage, current, and temperature values, but most of the existing work in the literature ignores this feature~\cite{vidal2020machine}. In this work, we found that the inclusion of the relative time feature significantly improved the accuracy of the models: e.g., RF accuracy (RMSE) of SOH estimation reaches 1.773\% when considering the relative time feature, while without the relative time, it increases to 116.20\%. 

\subsubsection{SOH Models Comparison}
Table~\ref{table:soh_comparison} clearly shows that the RF has the best performance with an RMSE = 1.773\%, followed by a slightly worse accuracy of LGB. The performance of the DNN is worse, with an RMSE = 7.114\%. The low accuracy of DNN is due to the limited size of our training dataset: machine learning generation techniques to do data augmentation for training the deep learning models may improve the accuracy. However, our experimental results demonstrate that simple machine learning techniques, e.g., RF and LGB, are completely capable of estimating the SOH.

\begin{table}[!ht]
    \centering
    \vspace{-0.5\baselineskip}
    \caption{Different SOH Models Performance} \label{table:soh_comparison}
    \begin{tabular}{|l|r|r|r|}
    \hline 
        \textbf{Model} & \textbf{Time (T + I)}  & \textbf{RMSE(\%)} & \textbf{MAE(\%)} \\ \hline
        \textbf{RF} & 13.704s + 0.47s & 1.773 & 0.603  \\ 
        \textbf{LGB} & 10.629s + 0.977s & 2.31 & 0.791  \\ 
        \textbf{DNN} & 399.582s + 1.439s & 7.114 & 1.696 \\ \hline
    \end{tabular}
\end{table}

Concerning time consumption (\textbf{T} represents training time, and \textbf{I} indicates inference time), the LGB method took the shortest time to train the model with only about 11 seconds, followed by the RF, that took approximately 14 seconds. Although the SOH model in our proposed digital twin is pre-trained offline and training time is not a decisive factor, too long training time can delay the deployment of the digital twin. For the comparison of SOH models in terms of time consumption, inference time is more critical than training time because this is computed in real-time in the cloud, but inference time of the three models is comparable (around 1s).

\subsection{SOC Estimation}

\subsubsection{Feature Selection}The same features used to train the SOH have been used to train the SOC models after being scaled using Min-Max normalization.

\subsubsection{SOC Models Comparison}
The SOC models investigated in this work show a similar performance as SOH models. Table~\ref{tab:soc_comparison_100} and ~\ref{tab:soc_comparison_70} show the performance of SOC models trained at the beginning (SOH=100\%) and when the SOH decreases to 70\% to conduct the SOC estimation at the same SOH level. RF and LGB show a high degree of accuracy, followed by DNN. Furthermore, the training time of DNN is relatively too long compared to the other two (approx. two orders of magnitude), and such a training time will weaken the overall performance of the digital twin, considering that the SOC model in our proposed digital twin architecture is constantly retrained and updated. Thus, the DNN method is proved to be less suitable than the others for the digital twin context. 

\begin{table}[!htbp]
    \centering
    \vspace{-0.5\baselineskip}
    \caption{SOC model performance at SOH = 100\%} \label{tab:soc_comparison_100}
    \begin{tabular}{|l|l|l|l|l|}
    \hline 
        \textbf{Model} & \textbf{Time (T + I)}  & \textbf{RMSE(\%)} & \textbf{MAE(\%)} & \textbf{Max Err.(\%)} \\ \hline
        \textbf{RF} &  0.825   + 0.034 & 0.038 & 0.132 & 1.442 \\
        \textbf{LGB} & 0.140   + 0.005 & 0.024 & 0.125 & 1.010\\ 
        \textbf{DNN} & 67.002 + 0.138 & 0.609 & 1.783 & 2.478 \\ \hline
    \end{tabular}
    \vspace{-1\baselineskip}
    \vspace{-1\baselineskip}
\end{table}

\begin{table}[!htbp]
    \centering
    \caption{SOC model performance at SOH = 70\%} \label{tab:soc_comparison_70}
    \begin{tabular}{|l|l|l|l|l|}
    \hline 
        \textbf{Model} & \textbf{Time (T + I)}  & \textbf{RMSE(\%)} & \textbf{MAE(\%)} & \textbf{Max Err.(\%)} \\ \hline
        \textbf{RF} & 0.622 + 0.017& 0.809 & 0.624 & 3.873 \\
        \textbf{LGB} & 0.097 + 0.005& 0.599 & 0.549 & 2.993 \\ 
        \textbf{DNN} & 29.97 + 0.068 & 2.775 & 1.380 & 5.071 \\ \hline
    \end{tabular}
\end{table}

\subsubsection{SOC Model Estimation at Different SOH Cases}

In this section, we illustrate why the SOC model needs to be retrained periodically by comparing the results of SOC estimation for the latest trained SOC model and the outdated SOC models. 

\begin{figure}[!htbp]
	\begin{center}
		\includegraphics[width=0.99\linewidth]{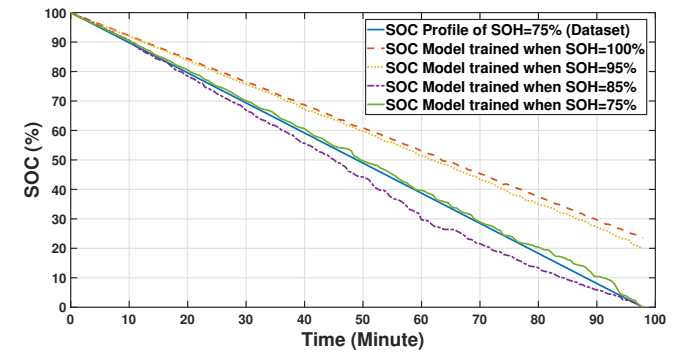}
		\caption{SOC discharge profiles comparison among different SOC Random Forest models trained in different SOH cases.}
		\label{fig:soc_retrain} 
		\vspace{-1\baselineskip}
	\end{center}
\end{figure}

Figure~\ref{fig:soc_retrain} summarizes the difference in SOC values estimated using the latest SOC RF model trained based on discharge data with SOH equal to 75\% and outdated SOC models trained based on SOH equal to 100\%, 95\%, 85\%, and 75\% when SOC estimation is performed at the time of SOH = 75\%.  
The solid blue dot line is the actual SOC discharge profile, and only the green circle line is closest to it (RMSE of these two lines is only 0.809\%, and MAE is 0.624\%), which is the value obtained from the SOC model trained using the discharge data at SOH=75\%. The SOC values obtained with the outdated SOC models are far from the correct values, and the older the model, the larger the error. This proves that it is necessary to retrain and update the SOC model periodically.

The first row in the Table~\ref{tab:soc_comparison_outdate} shows the errors of the SOC estimation at SOH=75\% when using the SOC model trained with SOH=100\%, corresponding to the difference between the orange star line and blue dot line shown in the Figure~\ref{fig:soc_retrain}. The other two models also show significant errors due to the use of too outdated SOC models. In particular, the MAE of DNN increases to 25.315\%. Considering that the range of SOH values is 0\% to 100\%, and the normal SOH operating range of the battery is only 80\% to 100\%, an error of more than 20\% is entirely unacceptable.

\begin{table}[!htbp]
    \centering
    \vspace{-1\baselineskip}
    \caption{soc model trained at SOH = 100\% used at SOH=75\% } \label{tab:soc_comparison_outdate}
    \begin{tabular}{|l|l|l|l|}
    \hline 
        \textbf{Model}  & \textbf{RMSE (\%)} & \textbf{MAE (\%)} & \textbf{Max Error (\%)} \\ \hline
        \textbf{RF}  & 184.450 & 11.731 & 23.516 \\
        \textbf{LGB}  & 188.917 & 11.858 & 23.666 \\ 
        \textbf{DNN}  & 859.818 & 25.315 & 54.067 \\ \hline
    \end{tabular}
    \vspace{-1\baselineskip}
\end{table}

\section{Conclusions}\label{sec:conclusions}
This work proposed a digital twin architecture to estimate battery SOH and SOC during the operation of an EV. The digital twin includes a model of SOH, built once for all on historical data, and a model of SOC that is periodically retrained to reflect the internal aging dynamics of the EV battery. The work focused on proving the impact of the proposed models on the relevance of preserving the relationship between SOC and SOH, and of adjusting models to the evolving conditions of a specific EV. The proposed solution has been exemplified on a public dataset to highlight the motivation experimentally and to prove its effectiveness in terms of extensibility to a variety of machine learning data-driven models and compatibility with onboard execution during EV operation (maximum inference times in the order of 1s, and retraining time for the SOC model lower than 1s). Future work will discuss the infrastructure supporting these models, including edge and cloud collaborative framework considerations. 

\bibliographystyle{IEEEtran}
\bibliography{biblio}

\end{document}